\documentclass[11pt,letterpaper]{article}
\usepackage{ijcnlp2017}
\usepackage{times}
\usepackage{latexsym}
\usepackage{graphicx}
\usepackage{siunitx}
\usepackage{amsmath}
\usepackage{algorithm}
\usepackage{amsmath}
\usepackage{pifont}
\usepackage{multirow}
\usepackage{enumitem}
\usepackage{CJKutf8}
\ijcnlpfinalcopy

\def\ijcnlppaperid{***}

\title{Auto Analysis of Customer Feedback using CNN and GRU Network}

\author{Deepak Gupta$^{\ast}$, Pabitra Lenka$^{\dagger}$, Harsimran Bedi$^{\ast}$, Asif Ekbal$^{\ast}$, Pushpak Bhattacharyya$^{\ast}$\\
$^{\ast}$Indian Institute of Technology Patna, India\\ $^{\dagger}$International Institute of Information Technology Bhubaneswar, India\\
		 {\tt $^{\ast}$\{deepak.pcs16, harsimran.mtcs16, asif, pb}\}{\tt @iitp.ac.in}\\
         {\tt $^{\dagger}$pabitra.lenka18@gmail.com}\\}

\date{}

\begin{document}

\maketitle

\begin{abstract}
Analyzing customer feedback is the best way to channelize the data into new marketing strategies that benefit entrepreneurs as well as customers. Therefore an automated system which can analyze the customer behavior is in great demand. Users may write feedbacks in any language, and hence mining appropriate information often becomes intractable. Especially in a traditional feature-based supervised model, it is difficult to build a generic system as one has to understand the concerned language for finding the relevant features.   
In order to overcome this, we propose deep Convolutional Neural Network (CNN) and Recurrent Neural Network (RNN) based approaches that do not require handcrafting of features. We evaluate these techniques for analyzing customer feedback sentences on four languages, namely English, French, Japanese and Spanish. 
Our empirical analysis shows that our models perform well in all the four languages on the setups of IJCNLP Shared Task on \textit{Customer Feedback Analysis}. Our model achieved the second rank in French, with an accuracy of 71.75\% and third ranks for all the other languages.
\end{abstract}

\section{Introduction}
Exploration and exploitation of customer feedbacks have become highly relevant and crucial for all the customer-centric business firms in today's world. Product manufacturers would like to know what their customers are liking or complaining about in order to improve their services or launch improved versions of products. Service providers would like to know how happy or unhappy a customer is with their service. 
According to a survey\footnote{\url{https://goo.gl/8KVwBh}}  96\% of unhappy customers do not complain but they advise 15\% of their friends to not have any business dealings with the particular firm.
With the huge amount of feedback data available, it is impossible to manually analyze each and every review. So there arises a need to automate this entire process to aid the business firms in customer feedback management. \\
\indent A customer review analysis can be associated with its sentiment polarity (`positive', `negative', `neutral' and `conflict') or with its interpretation (`request', `comment', `complaint'). 
There exists a significant number of works for sentiment classification \cite{pang2002thumbs, glorot2011domain, socher2013recursive, gupta2015pso, deepak2016opinion}, emotion classification \cite{yang2007emotion, li2009emotion, padgett1997representing} and customer review analysis \cite{yang2004online,mudambi2010makes, hu2004mining}.
However, the meaning of customer reviews (\textit{request, complaint, comment} etc.) remains a relatively not much explored area of research.\\
\indent Our present work deals with classifiyng a customer review into one of the six predefined categories. This can be trated as a document classification problem. 

The classes are \textit{comment, request, bug, complaint, meaningless} and \textit{undetermined}. The feedback classification is performed across four different languages, namely English (\textbf{EN}), French (\textbf{FR}), Japanese (\textbf{JP}) and Spanish (\textbf{ES}). In Table \ref{Example} we depict few instances of customer feedbacks with their label(s) in different languages. One of critical issues in traditional supervised model is to come up with a good set of features that could be effective in solving the problem. Hence it is challenging to build a generic model that could perform reasonably well acoross different domains and languages.  
In recent times, the emergence of deep learning methods have inspired researchers to develop solutions that do not require careful feature engineering. 
Deep Convolutional Neural Network (CNN) and Recurrent Neural Network (RNN) are two very popular deep learning techniques that have been successfully used in solving many sentence and document classification \cite{kim:2014:EMNLP2014, DBLP:journals/corr/XiaoC16} problems. 
We aim at developing a generic model 
that can be used across different languages and platforms for customer feedback analysis. \\
\indent
The remainder of our paper is structured as follows: Section \ref{relaed_work} offers the related literature survey for customer feedback analysis, where we discuss about the existing approaches. Section \ref{model} describes our two proposed approaches, one based on CNN and the other based on amalgamation of RNN with CNN. Section \ref{data_setup} provides the detailed information about the data set used in the experiment and the experimental setup. Results, analysis and discussion are elucidated in Section \ref{result_discussion}. We put forward the future work and conclude the paper with Section \ref{futureWork_conclusion}. The source code of our system can be found here.\footnote{\url{https://github.com/pabitralenka/Customer-Feedback-Analysis}}
\begin{table*}[!ht]
\begin{center}
\resizebox{\linewidth}{!}{%
\begin{tabular}{ c  c }
\hline
\textbf{Feedback} & \textbf{Class(es)} \\
\hline
nouveau bug : le rayon led anti yeux rouges se declanche intempestivement après avoir envoyé la photo ! & bug \\
\begin{CJK}{UTF8}{min}あと、タイムラインで他の人がお気に入りの記事を表示しないように設定できるようにしたい
\end{CJK} & request \\
Saw advertisements through an Internet travel booking site for hotel and zoo tickets in a package deal. & comment \\
La decoración en el hotel es excelente y las habitaciones tienen un toque chic.	& comment \\
it is fast, but the controls are lousy, plus it keeps installing on my desktop shortcuts to place I don't want.	& complaint \\
Mi pareja y yo hicimos una escapada romántica a Barcelona de cuatro días. & meaningless \\
Pour moi et avec modesties d'éloges, nero multimedia me rassure en ce sens que je peux:	& undetermined \\
\begin{CJK}{UTF8}{min}編集で付けようとしても、どうしてもその人のだけ消えてしまう	\end{CJK} & bug, comment \\ \hline
\end{tabular}%
}
\end{center}
\caption{\label{Example} Some instances of customer feedback in different languages annotated with their class(es)}
\end{table*}

\section{Related Work}\label{relaed_work}
Analysis of customer feedback has been of significant interest to many companies over the years. Given the large amount of feedbacks available, interesting trends and opinions among the customers can be investigated for many purposes. In this section, we discuss the related literature in the analysis of customer feedbacks.  \\
\indent \citet{Bentley2016GivingVT} developed a Office customer voice (OCV) system that classifies customer feedback on Microsoft Office products into known issues. The classification algorithm is built on logistic regression classifiers of the Python sci-kit framework. They have also employed a custom clustering algorithm along with topic modeling to identify new issues from the feedback data. A domain specific approach described in ~\cite{Potharaju:2013:JJT:2482626.2482640} infers problems, activities and actions from network trouble tickets. The authors developed a domain specific knowledge base and an ontology model using pattern mining and statistical Natural Language Processing (NLP) and used it for the inference. 
\citet{DBLP:journals/rcs/BrunH13} offers a pattern to extract suggestions for improvements from user reviews. They combine linguistic knowledge with an opinion mining system to extract the suggestive expressions from a review. The presence of a suggestion indicates that the user is not completely satisfied with the product. \\
\indent Over the years, many companies have developed feedback management systems to help other companies gain useful insight from the customer feedback data.  Customer satisfaction survey done by Freshdesk \footnote{\url{https://freshdesk.com/}} defines metrics for measuring customer satisfaction. They use a five class categorization (`positive', `neutral', `negative', `answered' and `unanswered'), thereby combining sentiment and responsiveness. Survey Monkey\footnote{\url{https://www.surveymonkey.com/}} also facilitates us to create survey forms and analyze them for customer satisfaction. It also uses a five-class categorization (`excellent', `good', `average', `fair' and `poor'), a commonly used rating mechanism. Customer Complaints Management\footnote{\url{http://www.newgensoft.com/solutions/cross-industry-solutions/customer-complaints-management/}} provides services and softwares to help business firms to manage and retain their customers. \\
Identifying a category for customer feedback requires deep semantic analysis of the lexicons to identify the emotions expressed. Authors in ~\cite{asher2009appraisal} have provided detailed annotation guidelines for opinion expressions where each opinion lies in the four top level categories of \textit{`Reporting', `Judgement', `Advise' and `Sentiment'}. \\
With the advent of deep learning, in recent years there has been a phenomenal growth in the use of neural network models for text analysis.
\citet{Yin:2016:RRY:2939672.2939677} have provided practical and effective comprehensive relevance solutions in Yahoo search engine. They have designed ranking functions, semantic matching features and query rewriting techniques for base relevance. Their network model incorporates a deep neural network and it is tested with the commercial Yahoo search engine. \\
\indent In this shared task the organizers have introduced a customer feedback analysis model where the task is to classify a feedback into one of the six categories. 
In following section we describe our proposed deep learning based classification model for customer feedback analysis.

\begin{figure*}[!ht]
\includegraphics[width=\textwidth]{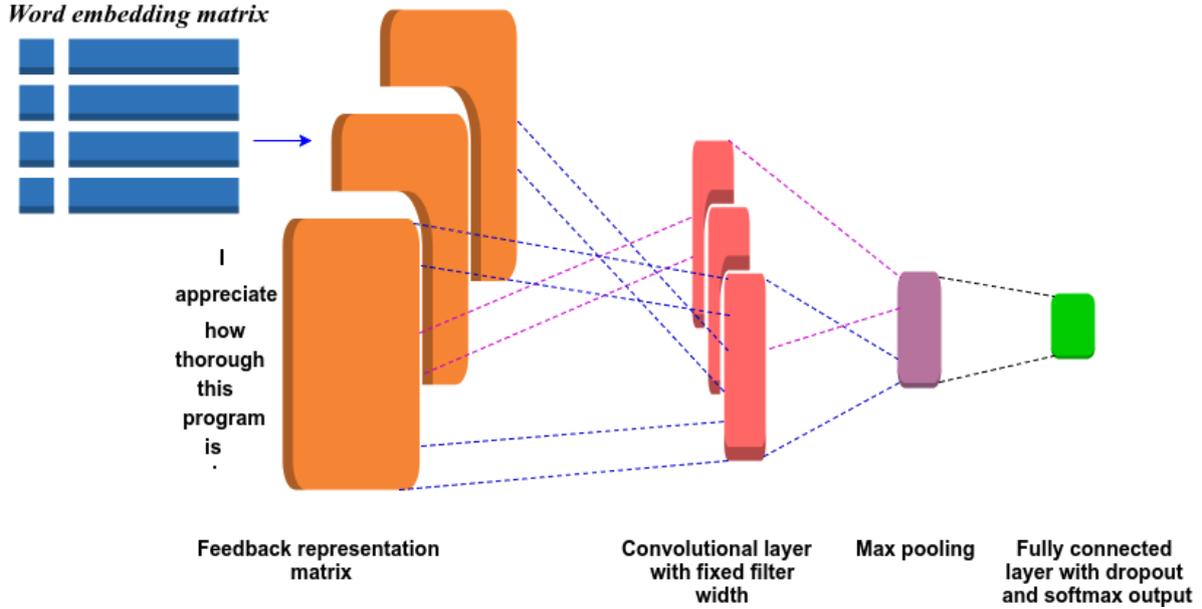}
\caption{\label{CNN image}Convolution neural network based feedback classification model}
\end{figure*}

\section{Network Architecture for Feedback Classification}\label{model}
In this section we describe our proposed neural network architecture for feedback classification. We propose two variants, the first one is convolution operation inspired CNN and the second one is the amalgamation of CNN with RNN. 
\subsection{Feedback classification using CNN} \label{model-1}
In this model a feedback sentence is subjected to CNN and the model predicts the most probable feedback class along with the confidence (probability) score. The model architecture is depicted in Fig \ref{CNN image}. The input and output of the model are as follows:\\
{\bf INPUT:} A feedback $F$, labeled with any of the six classes: (\textit{comment, request, bug, complaint, meaningless,} and \textit{undetermined})\\
{\bf OUTPUT:} Class(es) of the corresponding $F$ \\
Our model uses similar network architectures used by ~\citet{kim:2014:EMNLP2014} for performing the feedback classification task. We depict our model architecture in Figure \ref{CNN image}. The architecture of a typical CNN is composed of a stack of distinct layers where each layer performs a specific function of transforming its input into a useful representation. A CNN comprises of {\em sentence representation}, one or more {\em convolutional layers} often interweaved with {\em pooling layer} (max-pooling being extremely popular), followed by {\em fully-connected layer} leading into a $softmax$ classifier. The components of CNN are described as follows: \\

\subsubsection{Feedback Sentence Representation}
As CNNs deal with fixed length inputs, we ensure that every input feedback sentence has the same length. To achieve this, input feedback sentences are padded according to the need. Each feedback sentence is padded to the maximum sentence length\footnote{maximum feedback sentence length: EN=116, FR=73, JP=7 and ES=92}. Padding of feedback sentences to the same length is useful because it allows us to efficiently batch our data while training. Let a feedback sentence {\em F} consisting of `n' words be the input to our model such that $x = [x_1, x_2, \ldots x_n]$ where $x_i$ is the $i$\textsuperscript{th} word in the feedback sentence. Each token $x_i \in F$ is represented by its distributed representation $p_i \in R\textsuperscript{k}$ which is the $k$-dimensional word vector. The distributed representation $p$ is looked up into the word embedding matrix $W$ which is initialized either by a random process or by some pre-trained word embeddings like \textit{Word2Vec} ~\cite{NIPS2013_5021} or \textit{GloVe} ~\cite{pennington2014glove}. We then concatenate (row wise) the distributed representation $p_i$ for every $i$\textsuperscript{th} token in the feedback $F$ and build the feedback sentence representation matrix. The feedback sentence representation matrix $p_{1:n}$ can be represented as: 
\begin{equation}
p_{1:n} = p_1 \oplus p_2 \oplus \dots \oplus p_n
\end{equation}
where $\oplus$ is the concatenation operator. Each row of the sentence representation matrix corresponds to the word vector representation of each token. The result of the embedding operation yields a 3-dimensional tensor.

\subsubsection{Convolutional Layer}
The convolutional layer is the core building block of a CNN. The common patterns (n-grams) in the training data are extracted by applying the convolution operation. These patterns are then passed to the next hidden layer to extract more complex patterns, or directly fed to a standard classifier (usually a $softmax$ layer) to output the final prediction. The convolution operation is performed on the feedback representation matrix \textit{via} linear filters (feature detectors). Owing to the inherent sequential structure of the text data, we use filters with fixed \textit{width}. Then we simply vary the \textit{height} of the filter, i.e. the number of adjacent rows (tokens) considered together. Here the \textit{height} of the filter is the region size of the filter.

We consider a filter parameterized by the weight matrix $w$ with a region size $h$. Thereafter we denote the feedback representation matrix by $S \in R\textsuperscript{$n \times k$}$, where $k$ is the dimension of the word vector. The generated output $ out_i\in R\textsuperscript{$n-h+1$}$ of the convolutional operator is obtained by repeatedly applying the filter on sub-matrices of $S$:
\begin{equation}
out_i = w \cdot S[i : i + h - 1],
\end{equation}
The sub-matrix of $S$ from $i$\textsuperscript{th} row to $i+h-1$\textsuperscript{th} row is represented by $S[i : i + h - 1]$,
where $i = 1 \ldots n-h+1$ and $h$ is the height of the filter. A bias term $b \in R$ and an activation function $f$ to each $out_i$ is added which generates the {\em feature map} $c \in R\textsuperscript{$n-h+1$}$ for this filter where: 
\begin{equation}
c_i = f(out_i+b)
\end{equation}
But the dimensions of the {\em feature map} produced by each filter will differ as a function of the number of words in the feedback sentence and the filter region size. Thus we apply a pooling function over each {\em  feature map} to generate a vector of the fixed length.

\subsubsection{Pooling Layer}
The output of the convolutional layer is the input to the pooling layer. The primary utility of the pooling layer lies in progressively reducing the spatial dimensions of the intermediate representations. The operation performed by this layer is also called {\em down-sampling}, as there is a loss of information due to the reduction of dimensions. However, such a loss is beneficial for the network for two reasons:
\begin{enumerate}[nolistsep] 
\item Decreases the computational overhead of the network; and
\item Controls over-fitting.
\end{enumerate}
The pooling layer takes a sliding window or a certain region that is moved in stride across the input which transforms the values into representative values. There are several pooling operations in practice such as {\em max pooling, min pooling, average pooling} and {\em dynamic pooling}. We have applied the {\em max pooling} operation ~\cite{Collobert:2011:NLP:1953048.2078186} on the feature map which transforms the representation by taking the maximum value from the values observable in the window. Max pooling has been favored over others due to its better performance. It also provides a form of translation invariance and robustness to position. However,  ~\citet{DBLP:journals/corr/SpringenbergDBR14} have proposed to discard the pooling layer in the CNN architecture. 

\subsubsection{Fully-connected Layer}
Fully-connected layers are typically used in the final stages of the CNN to connect to the output layer. It looks at what high level features most strongly correlate to a particular class. The features generated from the pooling layer $p$ form the penultimate layer and are fed to a fully connected softmax layer to generate the classification. The $softmax$ classifier gives an intuitive output (normalized class probabilities) and also has a probabilistic interpretation. The output of the $softmax$ function is the probability distribution over tags ({\em comment, request, bug, complaint, meaningless,} and {\em undetermined}).
\begin{equation}
\begin{split}
P(c=i|F,p,z)& = softmax_i(p\textsuperscript{T}w_i+z_i)\\
&=\frac{\mathrm{e}^{p^{T}w_i+z_i}}{ \sum_{k=1}^{K}\mathrm{e}^{p^{T}w_k+z_k}}
\end{split}
\end{equation}
where $z_k$ and $w_k$ are the bias and weight vector of the $k$\textsuperscript{th} labels.

\subsection{Feedback Classification using CNN coupled with RNN}\label{model-2}
We propose a second method 
for feedback classification that combines both CNN and RNN. The typical architecture of these combinations is composed of a {\em convolutional feature extractor} applied on the input, then a {\em recurrent network} on top of the CNN’s output, then an optional {\em fully connected layer} is added to RNN’s output and finally fed into the {\em softmax layer}. We use convolutional layer as it learns to extract higher-level features that are invariant to local translation. By stacking up multiple convolutional layers, the network can extract higher-level, abstract, (locally) translation invariant features from the input sequence. 

Apart from this advantage, it is noticed that several layers of convolution are required to capture long-term dependencies, due to the locality of the convolution and pooling. In order to capture long-term dependencies even when there is only a single layer present in the network, the recurrent layer comes handy. However,
the recurrent layer increases the computational overhead due to its linearly growing computational complexity with respect to the length of the input sequence. So we have used a combination of convolutional and recurrent layers in a single model to ensure that it can capture long-term dependencies from the input more efficiently for the feedback classification task. Our model is similar to the one proposed in ~\citet{DBLP:journals/corr/XiaoC16}. They have used LSTM unit, however we have employed GRU \cite{cho-EtAl:2014:EMNLP2014}. 
\begin{equation}
\begin{split}
\nonumber
\mathbf{z}_{i}&=\sigma (\mathbf{W}_{z}c_{i}+\mathbf{V}_{z}\mathbf{h}_{i-1} + \mathbf{b}_{z} )  \\
\mathbf{r}_{i}&=\sigma (\mathbf{W}_{r}c_{i}+\mathbf{V}_{r}\mathbf{h}_{i-1} + \mathbf{b}_{r} )  \\
\mathbf{c}_{i} &=tanh (\mathbf{W}c_{i}+\mathbf{V} (\mathbf{r}_{i} \odot \mathbf{h}_{i-1}) + \mathbf{b}) \\
\mathbf{h_i}&=z_{i} \odot \mathbf{h}_{i-1} +  (1- \mathbf{z}_{i}) \odot \mathbf{c}_{i}
\end{split}
\end{equation}
where $\mathbf{z}_{i}$, $\mathbf{r}_{i}$ and $\mathbf{c}_{i}$ are update gate, reset get and new memory content, respectively. $c_{i}$ is the convolution output at time $t$. We take the last hidden states of both directions and
concatenate them to form a fixed-dimensional vector, which are later fed into the next layer.

\section{Datasets and Experimental Setup} \label{data_setup}
\subsection{Datasets}
The data sets used in our experiments are provided by the organizers of the shared task on \textit{Customer Feedback Analysis} of IJCNLP-2017. Data sets consist of representative real world samples of customer feedback from Microsoft Office customers in four languages, namely \textit{English, French, Japanese} and \textit{Spanish}. We obtain the English translations of test data of other three languages (\textit{French, Japanese} and \textit{Spanish}) which were translated using Google translate\footnote{\url{https://translate.google.com/}}. Each feedback in the data is annotated with one or multiple tags from the set of six tags (\textit{comment, request, bug, complaint, meaningless} and \textit{undetermined}). We show the dataset statistics for each language in Table \ref{distribution}. 
\begin{table*}[ht]
\sisetup{round-mode=places,round-precision=3}
\begin{center}
\resizebox{\linewidth}{!}{%
\begin{tabular}{ c | c  c  c  c  c  c  c | c  c  c  c  c  c  c | c  c  c  c  c  c  c }
\hline
 & \multicolumn{7}{ c |}{\textbf{Training}}  & \multicolumn{7}{ c |}{\textbf{Development}} & \multicolumn{7}{ c }{\textbf{Test}}\\
\cline{2-22}
\textbf{Language} & \textbf{CO} & \textbf{CP} & \textbf{RQ} & \textbf{BG} & \textbf{ME} & \textbf{UD} & \textbf{Total} & \textbf{CO} & \textbf{CP} & \textbf{RQ} & \textbf{BG} & \textbf{ME} & \textbf{UD} & \textbf{Total} & \textbf{CO} & \textbf{CP} & \textbf{RQ} & \textbf{BG} & \textbf{ME} & \textbf{UD} & \textbf{Total}\\
\hline

EN  & 1758 & 950 & 103 & 72 & 306 & 22 & \textbf{3211} & 276 & 146 & 19 & 20 & 48 & 3 & \textbf{512} & 285 & 145 & 13 & 10 & 62 & 4 & \textbf{519} \\
ES  & 1003 & 536 & 69 & 14 & 9 & 0 & \textbf{1631} & 244 & 39 & 12 & 5 & 1 & 0 & \textbf{301} & 229 & 53 & 14 & 2 & 1 & 0 & \textbf{299} \\
FR  & 1236 & 529 & 38 & 53 & 178 & 10 & \textbf{2044} & 256 & 112 & 6 & 8 & 36 & 1 &  \textbf{419} & 255 & 104 & 11 & 8 & 40 & 2 & \textbf{420}\\
JP  & 826 & 531 & 97 & 89 & 0 & 45 & \textbf{1588} & 142 & 73 & 22 & 18 & 0 & 9 & \textbf{264} & 170 & 94 & 26 & 14 & 0 & 9 & \textbf{313} \\ \hline
\textbf{Total}  & \textbf{4823} & \textbf{2546} & \textbf{307} & \textbf{228} & \textbf{493} & \textbf{77} & \textbf{8474} & \textbf{918} & \textbf{370} & \textbf{59} & \textbf{51} & \textbf{85} & \textbf{13} & \textbf{1496} & \textbf{939} & \textbf{396} & \textbf{64} & \textbf{34} & \textbf{103} & \textbf{15} & \textbf{1551} \\ \hline

\end{tabular}%
}
\end{center}
\caption{\label{distribution}Data set statistics of all the languages. Notations used are defined as follows, \textbf{CO:} \textit{comment}, \textbf{CP:} \textit{complaint}, \textbf{RQ:} \textit{request}, \textbf{BG:} \textit{bug}, \textbf{ME:} \textit{meaningless} and \textbf{UD:} \textit{undetermined},}
\end{table*}
\subsection{Data Preprocessing}
Some of the feedback sentences are annotated with multiple classes. Before feeding them to the network, we preprocessed the data by replicating the particular instance with all the possible classes. 


\subsection{Regularization}
In order to prevent the model from over-fitting, we employed a dropout regularization (set to 50\%) proposed by \citet{srivastava2014dropout} on the penultimate layer of the network. It ``drops out'' a random set of activations in the network. Dropout prevents feature co-adaptation by randomly setting some portion of hidden units to zero during the forward propagation when passing it to the softmax output layer in the end to perform classification. It also forces the network to be redundant i.e. it should be able to provide the correct classification or output for a specific input even if some of the activations are dropped out.

\subsection{Network Training and Hyper-parameters}

We have applied the rectified linear units (ReLu) ~\cite{Nair:2010:RLU:3104322.3104425} as the activation function in our experiment. We use the development data to fine-tune the hyper-parameters. In order to train the network, the stochastic gradient descent (SGD) over mini-batch is used and Backpropagation algorithm ~\cite{Hecht-Nielsen:1992:TBN:140639.140643} is used to compute the gradients in each learning iteration. We have not enforced L2 norm constraints on the weight vectors as ~\citet{DBLP:journals/corr/ZhangW15b} found that the constraints had a minimal effect on the end result. We have used cross-entropy loss as the loss function. The hyper-parameters of the best system in each language are listed in Table \ref{parameters}.  

\begin{table*}[!ht]
\sisetup{round-mode=places,round-precision=3}
\begin{center}
\begin{tabular}{ c  c | c  c  c | c  c  c }
\hline
& & \multicolumn{3}{| c |}{\textbf{CNN}}  & \multicolumn{3}{ c }{\textbf{CNN + RNN}}\\
\cline{3-8}
\textbf{Language} & \textbf{Tags} & \textbf{Precision} & \textbf{Recall} & \textbf{F1-Score} & \textbf{Precision} & \textbf{Recall} & \textbf{F1-Score} \\
\hline
Japanese  & comment & \num{0.563573883161512} & \num{0.964705882352941} & \num{0.711496746203904} & \num{0.569023569023569} & \num{0.994117647058823} & \num{0.723768736616702} \\
Japanese  & complaint & \num{0.166666666666666} & \num{0.0106382978723404} & \num{0.02} & \num{0.333333333333333} & \num{0.0106382978723404} & \num{0.0206185567010309} \\
French & comment & \num{0.789285714285714} & \num{0.866666666666666} & \num{0.826168224299065} & \num{0.784722222222222} & \num{0.886274509803921} & \num{0.832412523020257} \\
French & complaint & \num{0.630434782608695} & \num{0.557692307692307} & \num{0.591836734693877} & \num{0.602564102564102} & \num{0.451923076923076} & \num{0.516483516483516} \\
French & request & \num{-1} & \num{0} & \num{-1} & \num{1} & \num{0.0909090909090909} & \num{0.166666666666666} \\
French & meaningless & \num{0.576923076923076} & \num{0.375} & \num{0.454545454545454} & \num{0.53125} & \num{0.425} & \num{0.472222222222222} \\
Spanish & comment & \num{0.916666666666666} & \num{0.912663755458515} & \num{0.914660831509846} & \num{0.906382978723404} & \num{0.930131004366812} & \num{0.918103448275862} \\
Spanish & complaint & \num{0.597014925373134} & \num{0.754716981132075} & \num{0.666666666666666} & \num{0.655737704918032} & \num{0.754716981132075} & \num{0.701754385964912} \\
Spanish  & request & \num{1} & \num{0.285714285714285} & \num{0.444444444444444} & \num{1} & \num{0.214285714285714} & \num{0.352941176470588} \\
English  & comment & \num{0.826241134751773} & \num{0.817543859649122} & \num{0.821869488536155} & \num{0.712903225806451} & \num{0.775438596491228} & \num{0.742857142857142} \\
English  & complaint & \num{0.611428571428571} & \num{0.737931034482758} & \num{0.66875} & \num{0.5375} & \num{0.593103448275862} & \num{0.563934426229508} \\
English  & request & \num{1} & \num{0.0769230769230769} & \num{0.142857142857142} & \num{0.625} & \num{0.384615384615384} & \num{0.476190476190476} \\
English  & meaningless & \num{0.666666666666666} & \num{0.451612903225806} & \num{0.538461538461538} & \num{0.5} & \num{0.177419354838709} & \num{0.261904761904761} \\ \hline

\end{tabular}
\end{center}
\caption{\label{original} Performance results of the model \textit{with original test data} at the tags level. We did not provide the results of those tags which were not detected by our model.}
\end{table*}

\begin{table*}[!ht]
\sisetup{round-mode=places,round-precision=3}
\begin{center}
\begin{tabular}{ c  c | c  c  c | c  c  c }
\hline
& & \multicolumn{3}{| c |}{\textbf{CNN}}  & \multicolumn{3}{c }{\textbf{CNN + RNN}}\\
\cline{3-8}
\textbf{Language} & \textbf{Tags} & \textbf{Precision} & \textbf{Recall} & \textbf{F1-Score} & \textbf{Precision} & \textbf{Recall} & \textbf{F1-Score} \\
\hline

Japanese  & comment & \num{0.807692307692307} & \num{0.741176470588235} & \num{0.773006134969325} & \num{0.705882352941176} & \num{0.776470588235294} & \num{0.739495798319327} \\
Japanese & complaint & \num{0.571428571428571} & \num{0.76595744680851} & \num{0.654545454545454} & \num{0.551724137931034} & \num{0.51063829787234} & \num{0.530386740331491} \\
Japanese & request & -1 & 0 & -1 & \num{0.666666666666666} & \num{0.153846153846153} & \num{0.25} \\
Japanese  & bug & -1 & 0 & -1 & \num{0.5} & \num{0.0714285714285714} & \num{0.125} \\
French  & comment & \num{0.836501901140684} & \num{0.862745098039215} & \num{0.849420849420849} & \num{0.766323024054982} & \num{0.874509803921568} & \num{0.816849816849817} \\
French  & complaint & \num{0.679245283018867} & \num{0.692307692307692} & \num{0.685714285714285} & \num{0.651162790697674} & \num{0.538461538461538} & \num{0.589473684210526} \\
French  & request & \num{1.0} & \num{0.0909090909090909} & \num{0.166666666666666} & -1 & 0 & -1 \\
French  & meaningless & \num{0.433333333333333} & \num{0.325} & \num{0.371428571428571} & \num{0.434782608695652} & \num{0.25} & \num{0.317460317460317} \\
Spanish  & comment & \num{0.915178571428571} & \num{0.895196506550218} & \num{0.905077262693156} & \num{0.900862068965517} & \num{0.912663755458515} & \num{0.906724511930585} \\
Spanish  & complaint & \num{0.636363636363636} & \num{0.792452830188679} & \num{0.705882352941176} & \num{0.603448275862069} & \num{0.660377358490566} & \num{0.63063063063063} \\
Spanish  & request & -1 & 0 & -1 & \num{0.5} & \num{0.0714285714285714} & \num{0.125} \\ \hline

\end{tabular}
\end{center}
\caption{\label{translated} Performance results of the model with \textit{English translated test data} at the tags level.  We did not provide the results of those tags which were not detected by our model.}
\end{table*}

\subsection{Experiments}\label{experiment}
We conduct experiments in two different ways: CNN based and CNN+RNN based. Further we perform the experiments with \textit{original test data} and \textit{English translated test data}. In each setting we experiment with all the four languages\footnote{In the ``English translated test data'' setting, the experiments were performed with three languages (FR: French, JP: Japanese and ES: Spanish )}. Through the experimental results we wanted to establish the fact that whether a simple machine translation would work or there is a need of native tools for the other languages. 
The following are the descriptions of our submission in the shared task.
\begin{table*}[ht]
\begin{center}
\resizebox{\linewidth}{!}{%
\begin{tabular}{ c  c  c  c  c }
\hline
\textbf{Error Type} & \textbf{Language} & \textbf{Feedback} & \textbf{Reference} & \textbf{Predicted}\\
\hline
Ambiguous & EN & Make a paid version so we don't have to deal with the ads & request & complaint \\ 
Ambiguous & ES & La verdad, ir, ir, no va mal. & comment & complaint \\
Ambiguous & FR & \begin{tabular}[c]{@{}l@{}}Une bouilloire et du thé et du café dans la chAmbiguousre\\ (ainsi que sucre et lait). La salle de bain était grande.\end{tabular} & comment & complaint \\
Ambiguous & JP & \begin{CJK}{UTF8}{min}その後大浴場でサンセットを見てゆっくり入浴\end{CJK} & comment & complaint \\
Missing Target Entity & EN & Work with any type of PDF & comment & meaningless\\
Missing Target Entity & ES & El agua salpicaba el suelo del baño. & complaint & comment\\
Missing Target Entity & FR & de la grosse arnaque ! & complaint & meaningless \\
Missing Target Entity & JP & \begin{CJK}{UTF8}{min}英語の勉強にもなりそう \end{CJK} & comment & meaningless\\
Too short & EN & I gave up. & comment & complaint  \\
Too short & ES & Decepcionante & complaint & meaningless \\
Too short & FR & Brunch en famille & meaningless & comment \\
Too short & JP & \begin{CJK}{UTF8}{min} 使えん \end{CJK} & complaint & meaningless \\ \hline
\end{tabular}%
}
\end{center}
\caption{\label{error} Some of the feedback instances from different languages where our model failed to predict the correct tags. }
\end{table*}
\begin{table*}[ht]
\centering
\resizebox{0.7\linewidth}{!}{%
\begin{tabular}{ c  c  c  c  c }
\hline
\textbf{Parameter Name}                         & \textbf{EN} & \textbf{ES} & \textbf{FR} & \textbf{JP} \\ \hline
Embeddings                  & Pre-trained      & Random        & Pre-trained       & Pre-trained       \\
Maximum epochs                                  & 100         & 200         & 100         & 100         \\
 Mini batch size                                 & 64          & 64          & 64          & 64          \\
Number of filters              & 128         & 128         & 128         & 128         \\
Filter window sizes                             & 3,4,5       & 3,4,5       & 3,4,5       & 3,4,5       \\
Dimensionality of word embedding           & 300         & 300         & 300         & 300         \\

Dropout keep probability                        & 0.5         & 0.5         & 0.5         & 0.5         \\
                                        Hidden unit size (CNN+RNN)     & -           & 300         & -           & -           \\

 \hline
\end{tabular}%
}
\caption{\label{parameters} Network hyper-parameters for the best system (ref: Table \ref{comparison}) in each language}
\end{table*}
\begin{enumerate}
\item \textbf{CNN:} The results obtained from the CNN model described in Section \ref{model-1}.
\begin{enumerate}
\item \textbf{With original test data:} We train the CNN model using the dataset provided for training and used the respective test data to obtain the results. This setting of experiments are employed on all four languages (EN, FR, JP and ES).
\item \textbf{English translated test data:}  We train the CNN model using the English training dataset and use the English translated test data of other languages (FR, JP and ES) to obtain the results.
\end{enumerate}
\item \textbf{CNN+RNN:} The results obtained from the CNN+RNN model described in Section \ref{model-2}.
\begin{enumerate}
\item \textbf{With original test data:} Similar to CNN we train the CNN+RNN model using the dataset provided for training and use the respective test data to obtain the results. This setting of experiments are employed on all the four languages (EN, FR, JP and ES).
\item \textbf{English translated test data:}  Again similar to CNN, we train the CNN+RNN model using the English training dataset and use the English translated test data of other languages (FR, JP and ES) for the evaluation.
\end{enumerate}
\end{enumerate}
We use the pre-trained Google word embedding\footnote{\url{https://code.google.com/archive/p/word2vec/}} to initialize the word embedding matrix for English. The word embedding matrix for other three languages are initialized randomly\footnote{due to some computational issues, we were unable to use pre-trained embeddings of other languages.}.  
\section{Results and Discussions}\label{result_discussion}
We have submitted our results using both the models as discussed in Section \ref{experiment}. Table \ref{original} summarizes the performance of both the models \textit{with original test data}. Table \ref{translated} summarizes the performance of both the models on \textit{English translated test data}.
Table \ref{comparison} shows the exact accuracy comparison with the models which  achieve the best accuracy as compared to our model and also with the baseline scores. {Our systems easily predicted the true labels of sentences which had either positive connotation words like ``great", ``pleasant", ``nice", ``good", etc or negative connotation words like ``not", ``slow", ``unable", ``horrible", etc and classified them into \textit{comment} and \textit{complaint} classes respectively. The negative connotation words also appeared in the feedback sentences of \textit{bug} class. But owing to the larger amount of training data in the \textit{complaint} class as compared to the \textit{bug} class, the negative connotation words appeared significantly in the \textit{complaint} class. As a result, our systems had difficulty in predicting the true labels for the feedback sentences associated with the \textit{bug} class}. Our systems were unable to detect some tags due to the class imbalance problem in the training as well as test data. 
The scores of our systems could have been much better, provided that we should have more labeled training data. The system performance can be improved by the language specific pre-trained word embeddings.
\begin{table}[!ht]
\sisetup{round-mode=places,round-precision=2}
\begin{center}
\resizebox{\linewidth}{!}{%
\begin{tabular}{ c  c  c  c }
\hline
& \multicolumn{3}{ c }{\textbf{ Accuracy}}  \\ 
\cline{2-4}
\textbf{Language} & \textbf{Best System} & \textbf{Our Best System} & \textbf{Best Baseline} \\
\hline

EN & \num{71.00}\% & \num{70.00}\% (CNN)  & \num{48.8}\% \\
ES & \num{88.628762541806}\% & \num{85.6187290969899}\% (CNN+RNN)  & \num{77.257525083612}\% \\
FR & \num{73.75}\% & \num{71.75}\% (CNN-Trans) & \num{54.75}\%  \\
JP & \num{75.00}\% & \num{63.00}\% (CNN-Trans)  & \num{56.67}\% \\ \hline

\end{tabular}%
}
\end{center}
\caption{\label{comparison} Performance comparison with the best system in the shared task and the best baselines (3-gram features based SVM classifier). \textbf{CNN-Trans}: CNN model with English translated test data}
\end{table}

\subsection{Error Analysis} \label{error_analysis}
We perform error analysis on the outputs of our best performing model. Our system failed to detect some of the true positive classes due to some inadequacy in the training data. Table \ref{error} provides some examples (from different languages) where our system fails to detect the correct tags.
We divide those inadequacy into three different categories:
\begin{itemize}
\item \textbf{Ambiguous Feedback}:
Ambiguous feedback sentences have several possible meanings or interpretations. Our system fails to comprehend such doubtful or uncertain nature of customer feedback. 
\item \textbf{Missing Target Entity in Feedback}:
We found some feedback which were pretty straight without having a particular subject entitled to it. {These type of feedback sentences fail to address about what is being referred to in the sentences. These sentences do not sound complete. Let's take an example : ``Work with any type of PDF". It does not specify any comprehensive meaning. So the questions like ``What will work?", ``What is being talked about?" are bound to come up when the feedback sentences have an unstated subject}. This, in turn, generates misclassification.
\item \textbf{Too Short Feedback}:
There are several shorter-length feedback sentences, which are generic and do not provide any good evidence. 
Sometimes, these type of feedback sentences fail to convey any proper meaning to the end user who deals with it. The systems also experience difficulty to correctly tag those feedback sentences.
\end{itemize}

\section{Future Work and Conclusion} \label{futureWork_conclusion}
Convolutional neural networks (CNN) and recurrent neural networks (RNN) are architecturally two different ways of processing dimensioned and ordered data. These model the way the human visual cortex works, and has been shown to work incredibly well for natural language modeling and a number of other tasks. In our work, we extensively made use of CNN and RNN (GRU) to perform classification of customer feedback sentences into six different categories. Our proposed model performed well for all the languages. We have performed thorough error analysis to understand where our system fails. 
We believe that the performance can be improved by employing pre-trained word embeddings of the individual languages. Future work would focus on investigating appropriate deep learning method for classifying the short feedback sentences.

\bibliography{ijcnlp2017}
\bibliographystyle{ijcnlp2017}

\end{document}